\documentclass[runningheads]{llncs}
\usepackage{graphicx}
\begin{document}
\title{Hindi to English: Transformer-Based Neural Machine Translation}
%
%
\author{Kavit Gangar \and
Hardik Ruparel \and
Shreyas Lele}
\authorrunning{K. Gangar et al.}
%
\institute{Veermata Jijabai Technological Institute, Mumbai, India \\
\email{kavitgangar34@gmail.com, hardikruparel14@gmail.com, shreyaslele2398@gmail.com}\\}
\maketitle              
\begin{abstract}
Machine Translation (MT) is one of the most prominent tasks in Natural Language Processing (NLP) which involves the automatic conversion of texts from one natural language to another while preserving its meaning and fluency. Although the research in machine translation has been going on since multiple decades, the newer approach of integrating deep learning techniques in natural language processing has led to significant improvements in the translation quality. In this paper, we have developed a Neural Machine Translation (NMT) system by training the Transformer model to translate texts from Indian Language Hindi to English. Hindi being a low resource language has made it difficult for neural networks to understand the language thereby leading to a slow growth in the development of neural machine translators. Thus, to address this gap, we implemented back-translation to augment the training data and for creating the vocabulary, we experimented with both word and subword level tokenization using Byte Pair Encoding (BPE) thereby ending up training the Transformer in 10 different configurations. This led us to achieve a state-of-the-art BLEU score of 24.53 on the test set of IIT Bombay English-Hindi Corpus in one of the configurations.

\end{abstract}
\section{Introduction}
Machine translation is one of the oldest tasks taken up by computer scientists and the development in this field has been going on for more than 60 years. The research in this field has made remarkable progress to develop translator systems to convert source language to target language while maintaining the contextuality and fluency. In earlier times, the translation was handled by statically replacing words with the words from the target language. This dictionary look-up led technique led to inarticulate translation and hence was made obsolete by Rule-Based Machine Translation (RBMT) \cite{sandeepNMT}. RBMT is a system based on linguistic information about the source and target languages derived from dictionaries and grammar including semantics and syntactic regularities of each language \cite{rbmtWiki}. With the absence of flexibility and scalability to incorporate new words and semantics and the requirement of human expertise to define numerous rules, rule-based machine translation systems could only achieve accuracy on a subset of languages. To overcome the issues of the RBMT system, a new approach called Statistical Machine Translation (SMT) was introduced. Instead of having rules determine the target sequence, SMT approaches leverage probability and statistics to determine the output sequence. This approach made it feasible to cover all types of language within the source and target language and to add new pairs. Most of these systems are based on Bayesian prediction and have phrases and sentences as the basic units of translation. The main issue faced by this approach is the requirement of colossal amounts of data, which is a huge problem for low resource languages.

Due to these prevailing issues, there is a demand to explore alternate methods for creating a smarter and more efficient translation system. The development of various deep learning techniques and the promising results shown by the combination of these techniques with NLP created a new approach called NMT. NMT's advantage lies in two facts that are its simplistic architecture and its ability to capture long dependencies in the sentence, thereby indicating its huge potential in emerging as a new trend of the mainstream 
\cite{YangSurveyNMT}. Conceptually speaking, NMT is a simple Deep Neural Network (DNN) that reads the entire source sentence and produces an output translation one word at a time. The reason why NMT systems are appealing is that they require minimal domain knowledge which makes it well-suited for any problem that can be formulated as mapping an input sequence to an output sequence. Also, the inherent nature of the neural networks to generalize any input implies that NMT systems will generalize to novel word phrases that are not present in the training set.

Moreover, almost all the languages in the world are continuously evolving with new words getting added, older words getting amended and new slangs getting introduced very frequently. Even though NMT systems generalizes the input data well, they still lack the ability to translate the rare words due to their fixed modest-size vocabulary which forces the NMT system to use \emph{unk} symbol for representing out-of-vocabulary (OOV) words \cite{RareWordProblem}. To tackle this issue, a subword tokenization technique called Byte Pair Encoding (BPE) was introduced. BPE divides the words such that the frequent sequence of letters is combined thereby forming a root word and affix. This approach alone handles the OOV words by merging the root word and the different combinations of affixes thereby creating the rare word \cite{BPE}.

In this paper, we present the experimental setup and the state-of-the-art results obtained after training the Transformer model on the IIT Bombay CFILT English-Hindi dataset of 1.5 million parallel records. The paper is organized as follows: section \ref{sec:RW} describes the motivation behind our work. Section \ref{sec:the_model}, describes the model that we have implemented. In section \ref{sec:experimental_setup} we present the details of the experimental setup for training the model. In section \ref{sec:results} we display a comparative analysis of the results obtained by training the model in different configurations. Finally, section \ref{sec:conclusion} concludes the paper.

\section{Motivation}
\label{sec:RW}
With the power of deep learning, Neural Machine Translation has arisen as the most powerful approach to perform the translation task. 

In \cite{vaswaniTransformer} a model called Transformer was introduced which uses the encoder-decoder approach where the encoder first converts the original input sequence into its latent representation in the form of hidden state vectors. The decoder then tries to predict the output sequence using this latent representation. RNN's and CNN's inherently handle sequences word-by-word sequentially which is an obstacle to parallelize. The Transformer achieves parallelization by replacing recurrence with attention and encoding the  position of the symbol in the sequence. It reduces the number of sequential operations needed to relate the two symbols from the input/output phrases to a constant number O(1). This is achieved by using the multi-head attention mechanism which allows to model the dependencies regardless of their distance in input or output sentence.

In \cite{improveNMTMonolingual}, they show that they can improve the machine translation quality of NMT systems by mixing monolingual target sentences into the training set. They investigate two different methods to fill the source side of monolingual training instances: using a dummy source sentence and using a source sentence obtained via backtranslation, which they call synthetic data and conclude that the latter gives better results. 

In \cite{NMTRareWords}, they investigate NMT models that operate on the level of subword units. Their main goal is to model open-vocabulary translation in the NMT network itself, without requiring a back-off model for rare words. In this paper, byte pair encoding has been adapted. BPE is a compression algorithm that is used for word segmentation. It facilitates the representation of an open vocabulary via a fixed-size vocabulary that contains variable-length character sequences, thus it serves as a highly suitable strategy of word segmentation for neural network models.

Motivated with the results obtained by the transformer for machine translation on various languages, we created a translation system that translates a Hindi sentence to English. Since we use a low resource Hindi language for which the amount of good quality parallel corpus is limited, we applied back-translation to increase the quantity of training data. To overcome the problem caused by out of vocabulary words we used BPE.

\section{NMT Model Architecture}
\label{sec:the_model}
\subsection{Structure}
\begin{figure}[!h]

\includegraphics[width=\textwidth, height=11cm ]{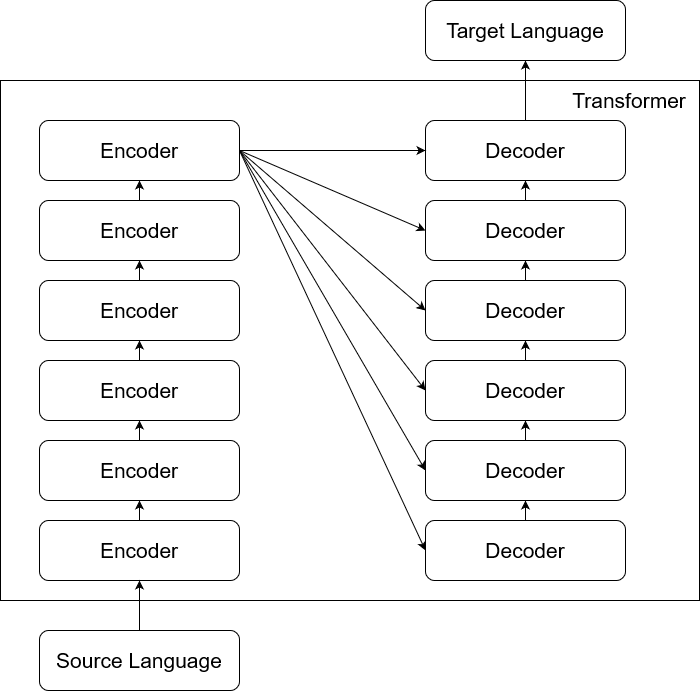}
\caption{Transformer Structure- Bird's-eye View} \label{transformer_structure}

\end{figure}

The Transformer model is the first NMT architecture that completely relies on the self-attention mechanism to calculate the representation of its input and output data without using recurrent neural networks (RNNs) or convolutional neural networks (CNNs) \cite{Goyal2019}. The Transformer model consists of an encoding unit and a decoding unit wherein each of these components consists of a stack of 6 layers of encoders and decoders respectively. (see Fig. \ref{transformer_structure}). 

\begin{figure}[!h]
\includegraphics[width=\textwidth, height=11cm]{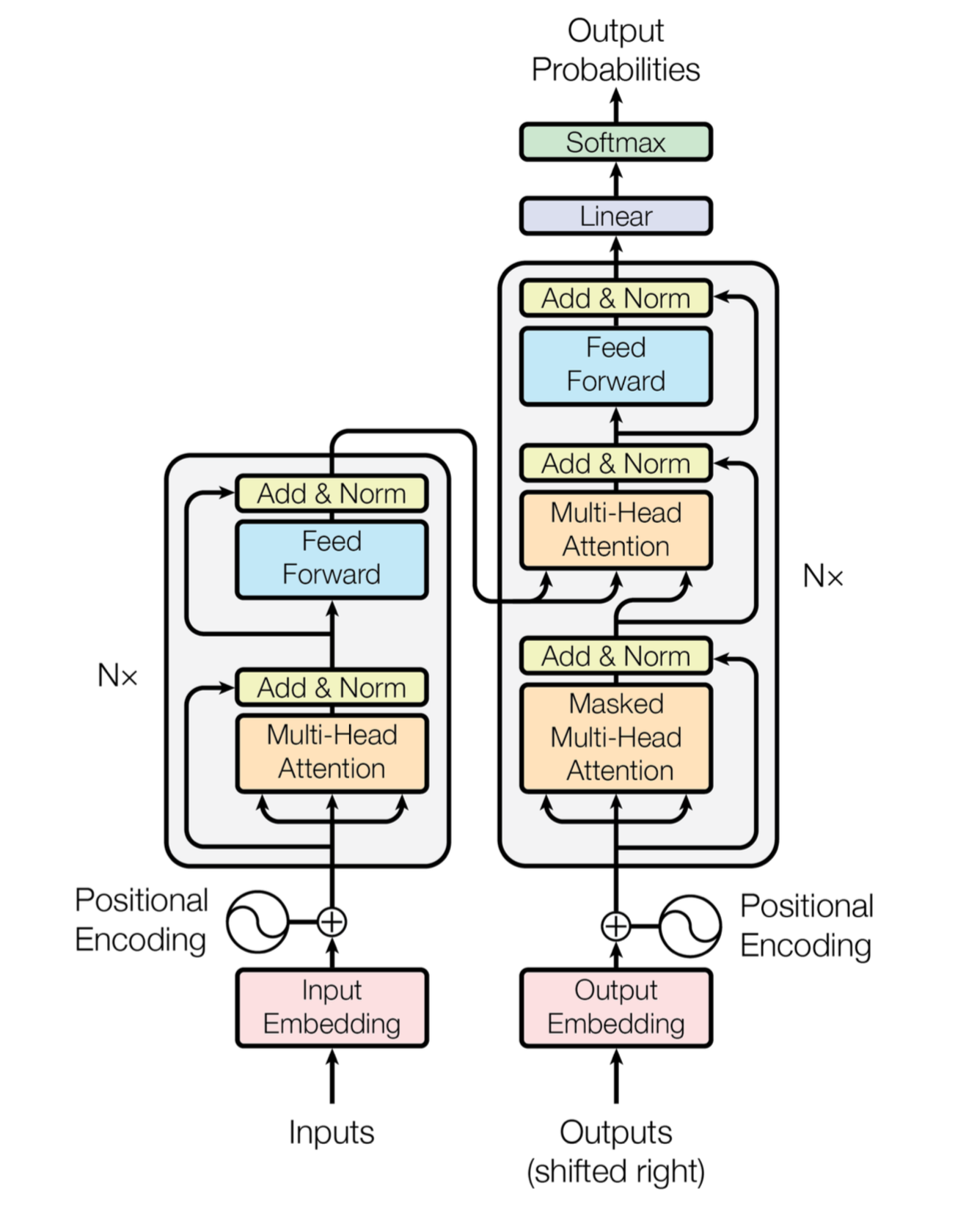}
\caption{The Transformer Architecture \cite{vaswaniTransformer}} \label{transformer_architecture}
\end{figure}

Each encoder layer consists of two sublayers. The first sublayer is the multi-head self-attention layer and the second sublayer is a position-wise fully connected feed forward network \cite{vaswaniTransformer}.  Each decoder layer in the decoding component consists of 3 sublayers. The function of the first two sublayers is the same as that in the encoder. However, the third sublayer performs multi-head attention mechanism over the output of the encoder stack (see Fig. \ref{transformer_architecture}).

\subsection{Working}
Before passing the input data to the encoder stack, the input data is first projected into an embedded vector space. To capture the notation, distance between different words and the order of the words in the input sequence, a positional embedding vector is added to the input vector. This intermediate vector is then fed to the first layer of the encoding component of the transformer. A multi-head self-attention is then computed on this intermediate embedded vector space. Multi-headed mechanism improves the performance of the attention-layer in two ways. First, it helps expand the model’s ability to focus on the words in different positions. Second, it gives the attention layer multiple representation subspaces, concatenates them and then projects linearly onto a space with initial dimensions \cite{illustratedTransformer}. The output of the self-attention layer is then passed onto a dense feed forward network which consists of two linear functions with RELU in between them. The output of this feed forward network is then passed on to another encoder layer stacked on top of it. All the encoder layers in the encoding component have the same functionality. Finally, the output of the encoding unit is then passed as an input to the decoding unit.

The decoder has similar functionality as that of the encoder. The output of the top encoder is converted into a set of attention vectors which is then used by each decoder in the decoding component. This helps the decoder focus on the appropriate position in the input sequence. The decoder predicts words one word at a time from left to right. Upon prediction of each word, it is again fed into the bottom decoder after converting it into an embedded vector space and adding a positional embedding vector. The decoder's self-attention mechanism works in a slightly different way than the encoder. In the decoder, the self-attention layer is only allowed to look at the words in earlier positions. This is done by masking the words at future positions to \emph{-inf}. Each decoder layer in the decoding component performs the same function. The output of the last decoder layer is then fed into a linear layer and softmax layer. The linear layer outputs a vector having a size equal to the size of the target language vocabulary. Each position in this output vector determines the score of the unique word. This vector of scores is then converted into probabilities by the softmax layer and the position with the highest probability is chosen, and the word associated with it is produced as the output for the particular time step.

\section{Experimental Setup}
\label{sec:experimental_setup}

\subsection{Dataset}
The fundamental requirement for assembling a machine translation system is the availability of parallel corpora of the source and the target language. In this paper, we trained our transformer model on the Hindi-English parallel corpus by the Center for Indian Language Technology (CFILT), IIT Bombay \cite{iitBData}. The training data consists of approximately 1.5 million texts from multiple disciplines while the development and the test set contains data from the news domain. Table \ref{dataset_metadata} provides the details about the number of sentences and the number of unique tokens in English and Hindi that are present in our chosen dataset. 

\begin{table}[b!]
\begin{center}
\caption{Metadata of the Dataset}\label{dataset_metadata}
\begin{tabular}{|c|c|c|c|}
\hline
\textbf{Dataset} &  \textbf{\# of Sentences} & \textbf{Unique Hindi Tokens} & \textbf{Unique English Tokens}\\
\hline
IITB Train & 1,267,502 & 421,050 & 242,910\\
IITB Dev & 483 & 2,479 & 2,405\\
IITB Test & 2,478 & 8,428 & 9,293\\
\hline
\end{tabular}
\end{center}
\end{table}

\subsection{Data Preprocessing}
Data preprocessing is an essential data mining technique that helps clean the data by removing the noise and outliers which can then directly be used by the model for training and testing purposes. Our preprocessing pipeline consists of 3 main steps viz. Data Cleaning, Removal of duplicates and Vocabulary creation. Each step is explained in detail below:

\subsubsection{Data Cleaning (Step 1)} In this step, we first removed the special characters, punctuation and noise characters from both the English and Hindi text corpus. After the elimination of all the noise characters, we removed the empty lines. The resulting text corpus was then converted into lower case and was then fed into the next step to remove the duplicates. 
\subsubsection{Removal of Duplicates (Step 2)} 
The cleaned and noise-free text corpus obtained as a result of the above step was then used to remove the duplicate records. This resulted in the creation of our training universe containing approximately 1.2 million unique parallel text corpus which was used for creating the vocabulary.

\subsubsection{Vocabulary Creation (Step 3)} Vocabulary creation is one of the most fundamental step in Neural Machine Translation. The coverage of the vocabulary is a major factor that drives the overall accuracy and the quality of the translation. If the vocabulary under-represents the training data universe, then predicted translation will contain many \emph{unk} tokens thereby reducing the BLEU score drastically. Thus creating a modest-size vocabulary that optimally represents the data is a challenging task. For creating the vocabulary for both Hindi and English language we implemented two approaches: word level tokenization and subword level tokenization. In the word level tokenization, we first extracted 50,000 most frequently used words from the training set and then added it to the vocabulary. While in the subword level tokenization, we used Byte Pair Encoding (BPE) for creating 50,000 subword tokens which were added in the vocabulary. The evaluation of the performance of our model on both word and subword level tokenization is presented in Section \ref{sec:results}.

\subsection{Back-Translation}
\label{sec:back_translation}
Hindi, being a low resource language as compared to its counterpart European languages has made the availability of data quite difficult. Many institutions around the world are creating larger and a more complete text corpus for the low resource languages. To tackle the lack of availability of Hindi-English parallel corpus, we implemented back-translation technique. Back-translation technique is used for augmenting the training data which leads to increasing the output accuracy of the translation. There is a plethora of monolingual English data available on the internet which can be used to generate text corpus of a low resource language. To generate the additional Hindi-English parallel text corpus, we first trained an English to Hindi machine translation system on our training data and then translated the 3 million WMT14 English monolingual data to generate the corresponding predicted Hindi text corpus. 

\begin{table}[h!]
\begin{center}
\caption{Training Data Universe: Batch-wise summary }\label{training_universe}
\begin{tabular}{|c|c|c|}
\hline
\textbf{Batch Number} &  \ \textbf{\# of Back-translated records added}  & \ \textbf{Total Records} \\
\hline
Batch 1 & 0.5 million & 1.7 million \\
Batch 2 & 1.5 million & 2.7 million \\
Batch 3 & 2.5 million  & 3.7 million \\
Batch 4 & 3 million & 4.2 million \\

\hline
\end{tabular}
\end{center}
\end{table}

To observe the effect of back-translation, we have divided the 3 million back-translated parallel records in 4 batches. We cumulatively add the back-translated records to the original training data in each of these batches. The first batch contains the 0.5 million back-translated records along with the 1.2 million original training data. In the same way we add an additional 1 million, 1 million and 0.5 million in the second, third and fourth batch respectively. Table \ref{training_universe} summarizes the training data universe for each batch.

\subsection{Training Details}
\label{sec:training_details}
After the data preparation and segregation into batches, we trained our transformer model using Opennmt-tf toolkit \cite{opennmtTool}. For training the model, we have used the NVIDIA Tesla K80 GPU provided by Google Colab \cite{colab}. For our transformer model, we used the default 6 layers setting in both encoder and decoder each of which contains 512 hidden units. We used the default embedding size of 512 along with 8 headed attention. We configured the batch size to be equal to 64 records and the effective batch size which is defined as the number of training examples consumed in one step to be equal to 384. We optimized the model parameters using the LazyAdam optimizer. The model was trained on 10 different configurations 5 each for word and subword level tokenization till convergence or till 70,000 steps at max (hard stop). The GPU run-time provided on Google Colab resulted in a training duration of approximately 20-24 hours for each configuration.

\section{Results}
\label{sec:results}
We evaluate the quality of translation of our  model on the test set using the Bilingual Evaluation Understudy (BLEU) score \cite{BLEU} and the Rank-based Intuitive Bilingual Evaluation (RIBES) score \cite{RIBES}. For depicting the performance of subword level tokenization, we have divided the test set into 2 subsets. The first set (Set-1) consists of sentences whose words are present in the vocabulary generated from word level tokenization. This set consists of 1694 sentences. The second set (Set-2) is the complete test set consisting of 2478 sentences.

\begin{table}[b!]
\begin{center}
\caption{Results of Word Level Tokenization (Set-1) }\label{result_word_level}
\begin{tabular}{|c|c|c|c|}
\hline
\textbf{Model ID} & \textbf{Model} &  \ \textbf{BLEU} & \ \textbf{RIBES} \\
\hline
1 & Transformer & 18.76 & 0.699708 \\
2 & Transformer with Batch 1 & 22.55 & 0.730440 \\
3 & Transformer with  Batch 2 & 23.95  & 0.735804 \\
4 & Transformer with Batch 3 & 24.79 & 0.741369 \\
5 & Transformer with Batch 4 & 24.68 & 0.740567 \\
\hline
\end{tabular}
\end{center}
\end{table}

In Table \ref{result_word_level}, after adding the first batch of 0.5M parallel back-translated records with the original training data, the BLEU score increased by 3.79 and with the subsequent addition of other 2 batches the BLEU and the RIBES score reached a maximum of 24.79 and 0.741 respectively. However, when the 4th batch of 0.5M back-translated records was added with the previous batches, the scores decreased by a small margin indicating convergence with respect to the addition of back-translated data.

\begin{table}[t!]
\begin{center}
\caption{Results of Subword Level Tokenization (Set-1) }\label{result_subword_level}
\begin{tabular}{|c|c|c|c|}
\hline
\textbf{Model ID} & \textbf{Model} &  \ \textbf{BLEU} & \ \textbf{RIBES} \\
\hline
6 & Transformer & 19.10 & 0.695566 \\
7 & Transformer with Batch 1 & 23.98  & 0.733614  \\
8 & Transformer with Batch 2 & 25.44 & 0.737078 \\
9 & Transformer with Batch 3 & 25.87 & 0.739192 \\
10 & Transformer with Batch 4 & 25.74 & 0.742397 \\
\hline
\end{tabular}
\end{center}
\end{table}
Similar to the results obtained with word level tokenization, in Table \ref{result_subword_level}, after adding the first batch of back-translated records the  BLEU score increases by 4.78 and with the subsequent addition of other 2 batches the BLEU score reached a maximum of 25.87. After adding the 4th batch, the BLEU scored decreased by 0.13 however the RIBES score increased by 0.003.

When compared with word level tokenization, subword level tokenization achieves a better BLEU score which can be attributed to the fact that it has the advantage of not having an out-of-vocabulary case and also to learn better embeddings for rare words since rare words can enhance the learning from its subwords that occur in other words. This fact is further strengthened in Table 5 which shows the BLEU and RIBES score for the Transformer with Batch4 model using word and subword level tokenization on Set-2. The decrease in the BLEU and RIBES score as compared to Table \ref{result_word_level} and Table \ref{result_subword_level} is due to the fact that the Set-2 consists of additional sentences as compared to Set-1 which contain rare words that are not included in the vocabulary for word level tokenization. When subword level tokenization is used, the model performs reasonably well even in the presence of rare words which is not the case for word level tokenization. 

\begin{table}[t!]
\begin{center}
\caption{Results for the Transformer with Batch4 model on Set-2}\label{final_result}
\begin{tabular}{|c|c|c|c|}
\hline
\textbf{Model ID} & \textbf{Tokenization} &  \ \textbf{BLEU} & \ \textbf{RIBES} \\
\hline
5 & Word level & 21.22 & 0.728683 \\
10 & Subword level & 24.53  & 0.735781  \\

\hline
\end{tabular}
\end{center}
\end{table}

\section{Conclusion}
\label{sec:conclusion}
The transformer model has displayed promising results for neural machine translation involving low resource languages as well. We saw that after adding the back-translated records the performance was certainly improved, however when the amount of generated data increases beyond a certain level, there is no further improvement in the performance. Using a combination of the transformer model, back-translation technique and a subword tokenization method like BPE, we achieved a BLEU score of 24.53 which is the state-of-the-art on this dataset to the best of our knowledge. In the future, we can try to incorporate state-of-the-art Natural Language Processing models like BERT \cite{devlinBERT} into NMT to further improve the quality of translation.

%
%

\end{document}